# Observation Subset Selection as Local Compilation of Performance Profiles

**Yan Radovilsky**
Computer Science Dept.
Ben-Gurion University
84105 Beer-Sheva, Israel
yanr@cs.bgu.ac.il

**Solomon Eyal Shimony**
Computer Science Dept.
Ben-Gurion University
84105 Beer-Sheva, Israel
shimony@cs.bgu.ac.il

## Abstract

Deciding what to sense is a crucial task, made harder by dependencies and by a non-additive utility function. We develop approximation algorithms for selecting an optimal set of measurements, under a dependency structure modeled by a tree-shaped *Bayesian network* (BN).

Our approach is a generalization of composing *anytime algorithm* represented by *conditional performance profiles*. This is done by relaxing the *input monotonicity* assumption, and extending the *local compilation* technique to more general classes of *performance profiles* (PPs). We apply the extended scheme to selecting a subset of measurements for choosing a maximum expectation variable in a binary valued BN, and for minimizing the worst variance in a Gaussian BN.

## 1 Introduction

A typical diagnostic system consists of two types of variables: tests (observable) and hypotheses (unobservable), with statistical dependencies among variables. Each test, if performed, consumes resources (time or money), and provides a measurement of one or more test variables. After obtaining the values of the selected tests, the distribution of the model is updated. An objective function specifies a reward given to the system for the posterior distribution. The system should make its selection so as to optimize the objective function, a hard problem in the general case. *Observation subset selection* (OSS) is a restricted version of this problem, where all measurements must be selected in advance, prior to any observations. In this paper we develop approximation algorithms for some settings of the OSS problem for tree-shaped dependency structures.

To tackle this problem we present OSS as a variant of the following well-known meta-reasoning problem. In systems composed of several *computational components* (CCs), the meta-level controller should reason about allocation of available computational resources for different CCs in order to optimize the overall performance of the entire system. This task is usually referred to in the research literature as the *meta-level resource allocation* (MRA) problem (see for example [11]). The standard approach used to optimize the MRA task was proposed by S. Zilberstein [10, 11, 12], the technique of *local compilation* (LC). This technique is applied to individual CCs, represented in a form of *conditional performance profiles*, and generates the optimal *time allocation scheme* for the entire system (see Section 2). However, local compilation requires the *input monotonicity* assumption and is, therefore, restricted to deterministic *performance profiles* with scalar output quality.

In this paper we relax the *input monotonicity* assumption and extend the LC technique to more general classes of PPs. We then apply the extended approximation scheme (Section 4) to find an approximate solution to the OSS problem, under two settings, both for dependencies modeled as a tree-shaped BN: a) finding a maximum expectation variable in a binary valued BN (Section 4.1), and b) minimizing the worst variance in a Gaussian BN (Section 4.2).

## 2 Background

*Flexible computation* refers to procedures that allow a graceful trade-off to be made between the quality of results and allocation of costly resources, such as time, memory, or information [3]. Since time is usually the main computational resource, there are several alternative terms used for reference to *flexible computation* in the research literature: *continual computation* [4, 5], *anytime computation* [8], and *anytime algorithms* [1, 10, 12].

To predict the quality of the result which depends on the amount of allocated time, a statistical model called a *performance profile* (PP) is employed. The most simple version of such a PP called an *expected performance profile* (EPP), a mapping from consumed time to an expected result quality, $Q : \mathbf{T} \to \mathbf{Q}$. Given *anytime algorithm A* described by EPP $Q_A$ and time-dependent utility function $U : \mathbf{T} \times \mathbf{Q} \to \mathcal{R}$, optimal time allocation $t^*$ can be derived as follows:

$$t^* = \arg\max_t U(t, Q_A(t)) \qquad (1)$$

When an algorithm operates with some inexact inputs, the quality of its result may also strongly depend on the quality of the inputs. To address this point, a more flexible model, known as a *conditional performance profile* (CPP), is used. For example, an *anytime algorithm* with 2 inputs and one output can be represented by a CPP $Q : \mathbf{T} \times \mathbf{Q}^2 \to \mathbf{Q}$.

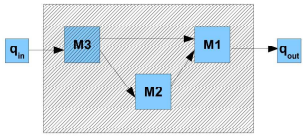

Figure 1: Composition graph

In a complex system, several CCs (or *anytime algorithms*) can be composed to achieve a required result. In this case, their PPs should be compiled in order to obtain an appropriate performance prediction for the entire system. Such a system is usually described in a graphical form by a *directed acyclic graph* (DAG), where each node corresponds to a single CC and is associated with an appropriate PP, while edges represent dependencies between input/output qualities of different CCs (Figure 1). This model is referred to as a *composition graph* (CG).

A *time allocation scheme* (TAS) $\hat{t} \in \mathbf{T}^n$ specifies allocation of time for each of $n$ elementary CCs of a composite system. Output quality of the whole system can be expressed as a function of a TAS in a form of *composite expression* (CE) $\phi : \mathbf{T}^n \times \mathbf{Q}^k \to \mathbf{Q}$, where $k$ is a number of external inputs. For example, the system in figure 1 has the following CE:

$$\phi(\hat{t}, q_{in}) = Q_1([\hat{t}]_1, Q_2([\hat{t}]_2, Q_3([\hat{t}]_3, q_{in})), Q_3([\hat{t}]_3, q_{in}))$$

The MRA problem for a *composite system* is to optimize a time-dependent utility function by selecting an appropriate TAS. Formally, given a *composite system* represented by CE $\phi$, external input quality $q_{in}$, and utility function $U$, the goal is to find TAS $\hat{t}^* \in \mathbf{T}^n$ which maximizes the utility:

$$\hat{t}^* = \arg\max_{\hat{t} \in \mathbf{T}^n} U(\|\hat{t}\|, \phi(\hat{t}, q_{in})), \qquad (2)$$

where operator $\|\cdot\|$ denotes a summation over all elements of its argument vector: $\|\hat{t}\| = \sum_i [\hat{t}]_i$.

An intuitive approach to tackle the MRA problem has been proposed in [10]. This approach, called *compilation of performance profiles*, involves construction of an appropriate PP for the entire system:

$$Q^c(t, q_{in}) = \max_{\hat{t}\in\mathbf{T}^n : \|\hat{t}\|=t} \phi(\hat{t}, q_{in}), \qquad (3)$$

Each entry in $Q^c$ is additionally associated with a corresponding TAS $\pi(t, q_{in}) = \hat{t}$ s.t. $\phi(\hat{t}, q_{in}) = Q^c(t, q_{in})$ and $\|\hat{t}\| = t$.

Once such a model is constructed, the optimal solution for the MRA problem can be derived from this model as $\hat{t}^* = \pi(t^*, q_{in})$, where $t^*$ is the optimal total time allocation computed using equation 1.

---

**Algorithm 1**: method `RLC` for in-tree shaped composite systems

**Input**  : $s$ - a node in a composition graph.
**Output**: Composed PP for the subtree rooted in $s$.
**if** *(Pred(s) = ∅)* **then**
    return $Q_s$;
**foreach** $s_i \in $ *Pred(s)* **do**
    Let $Q^c_{s_i} \leftarrow $ *RLC*$(s_i)$;
Let $L \leftarrow \{Q^c_{s_i} : s_i \in $ *Pred(s)*$\}$;
**return** *Compose*$(Q_s, L)$;

---

In general, the task of (global) *compilation* is computationally hard even for approximate solution [10]. However, for some restricted topologies (e.g. trees) S. Zilberstein [10] proposed an efficient algorithm based on the *local compilation* (LC) technique, summarized as RLC in Algorithm 1. In this technique a composite PP of each subtree is generated based on the CPP of its root and composed PPs of all its predecessors. Method *Compose* performs the basic composition operation, which results in the following composite PP:

$$Q^c_s(t, q_1, \ldots, q_k) = \qquad (4)$$
$$\max_{t_0,\ldots,t_k \in \mathbf{T}: \sum t_i = t} Q_s(t_0, Q^c_{s_1}(t_1, q_1), \ldots, Q^c_{s_k}(t_k, q_k))$$

where $s_1, \ldots, s_k$ are the predecessors of node $s$.

For systems, where all CCs are represented by deterministic CPPs, RLC is proved to produce the correct result (equals to one obtained by *global compilation*) when the following assumptions hold:

- A *tree-structured* CG - each node has only one output which serves as an input for one successor node except for the root node, whose output is the resulting system output.

- *Input monotonicity* (IM) - the utility function and all involved CPPs are non-decreasing functions of input quality, i.e. $\forall q \geq p : Q(t,q) \geq Q(t,p)$.

Such restrictive assumptions are required to enable independent compilation of each subtree followed by *greedy* selection of one local TAS for each total allocation without harming optimality of the resulting solution. However, in practice, the IM assumption may be too restrictive. For example, a diagnostic system with a dependency model represented by Gaussian BN and a utility function depending on posterior marginal precision of several hypothesis nodes, violates this assumption. In this work we propose an extension to the CPP model, which allows relaxing the IM assumption.

Recent work exists in another line directly related to the OSS task. In [7] the authors proved that the OSS problem is $NP^{PP}$-hard even for tree-structured BNs. They proposed a polynomial time algorithm, based on *dynamic programming*, which constructs an optimal solution to a version of OSS restricted to chain topology, exact observation, and additive reward. In [9] a similar technique was used beyond the *exact observation* assumption, and determined a theoretical bound for the worst-case loss in expected reward. Our work can be seen as an extention of the latter research. Another approximation method based on greedy selection of test nodes is applicable when the reward function exhibits property of *sub-modularity* [6].

## 3  Generalized Local Compilation

This section generalizes the notion of a *performance profile* to a *reachable performance profile*, and adapts the LC technique to handle the extended model.

**Definition 3.1** (Conditional performance). *Conditional performance (CP) of a CC (either elementary or composite) is a tuple $(\hat{t}, p, q)$, where $\hat{t}$ is a local TAS (t-component), p is a required input quality (p-component), and q is an expected output quality, obtained by the CC when applied to inputs of quality p with TAS $\hat{t}$ (q-component).*

Either $p$ or $q$ may be represented by scalars or by vectors, while $\hat{t}$ is assumed to be a complete assignment of time allocations to all CCs of the system.

**Definition 3.2** (Reachable performance profile). *Reachable performance profile (RPP) of a CC is a set of CPs achievable by this CC.*

The RPP model can be derived from a CPP as follows:

$$R_A = \{(\pi_A(t,p), p, Q_A(t,p))\}, \quad (5)$$

where $Q_A$ is a CPP, and $R_A$ is an appropriate RPP. The converse is not always possible, because a RPP can contain more than one output quality for the same pair of total time allocation and input quality (each with a different *local* TAS).

We further extend our model by allowing backward conditioning, which means that output quality of a CC may depend on output quality of its successor. The general form of a CP in this case is $(\hat{t}, (p_{suc}, p_{pred}), (q_{suc}, q_{pred}))$, where the p-component has two parts: $p_{suc}$ is a required output quality of the successor, and $p_{pred}$ is a vector of required output qualities of all the predecessors; the q-component has two parts as well: an output quality towards its successor ($q_{suc}$), and a vector of qualities towards all its predecessors ($q_{pred}$). Such a form of CPs is useful in applications for BNs, where posterior probability distribution of a node (and its local reward) depends on the messages coming from all its neighbors (details appear in section 4).

In order to adapt the RLC algorithm to the extended settings, we need to modify the *Compose* method. This method is now applied to a list of profiles in the RPP representation, and its output should be in the RPP form as well. Moreover, the greedy selection reflected in the max operator in equation 3 strongly relies on the IM assumption. Since this assumption fails in the RPP case, we propose another approach. The composition process comprises two parts: the first part (*Construct*) considers all combinations of input CPs, and collects appropriate resulting CPs.

$$Q = \{(\hat{t}_0 + \hat{t}_1 + \ldots + \hat{t}_k, p_0, q_0) : (\hat{t}_i, p_i, q_i) \in Q_{s_i}, \\ (\hat{t}_0, (p_0, q_1, \ldots, q_k), (q_0, p_1, \ldots, p_k)) \in Q_s\}. \quad (6)$$

The second part, called *Purge* is applied to the set Q. *Purge* exploits *domination* and *equivalence* among reachable CPs in order to filter out irrelevant CPs.[1] The idea of pruning irrelevant CPs is very similar to one used in the *Incremental Pruning* algorithm [2] for filtering irrelevant $\alpha$-vectors. The resulting RPP $Q^c$ keeps one representative for each CP in Q:

$$\forall a \in Q \ \exists b \in Q^c \text{s.t.} \ (b \succ a) \vee (b \approx a).$$

where $\succ$ and $\approx$ are *domination* and *equivalence* operators respectively. These operators can be defined based on partial orders within each performance component. Assuming only a natural partial order in the t-component we obtain:

$$(\hat{t}, p, q) \approx (\hat{t}', p', q') \Leftrightarrow (\|\hat{t}\| = \|\hat{t}'\|) \wedge (p \stackrel{P}{\approx} p') \wedge (q \stackrel{Q}{\approx} q')$$

and

$$(\hat{t}, p, q) \succ (\hat{t}', p', q') \Leftrightarrow (\|\hat{t}\| < \|\hat{t}'\|) \wedge (p \stackrel{P}{\approx} p') \wedge (q \stackrel{Q}{\approx} q')$$

---
[1] The filtering can be applied after *Construct* terminates, but it is usually more efficient to perform filtering on-the-fly.

where $\stackrel{P}{\approx}$ and $\stackrel{Q}{\approx}$ are equivalence operators defined over the $p$ and $q$ components, respectively.

## 4 Observation Selection in BNs.

In this section we describe how our framework can be applied to OSS in BNs. We use the following notation:

- $\mathbf{X} = \{X_i : 1 \le i \le n\}$ - set of all state variables;
- $\mathbf{X}_H \subseteq \mathbf{X}$ - set of hypothesis state variables;
- $\mathbf{X}_M \subseteq \mathbf{X}$ - set of measurable state variables;
- $\mathbf{Y} = \{Y_i : X_i \in \mathbf{X}_M\}$ - set of test variables;
- $N$ - BN over $\mathbf{X} \cup \mathbf{Y}$;
- $R : \Pr(\mathbf{X}_H) \to \mathcal{R}$ - reward function;
- $T(E) = \sum_{Y_i \in E} \tau_i$ - additive time consumption;
- $B$ - time budget (maximum time for observation).

**Definition 4.1** (OSS optimization problem). *The OSS optimization problem is: given a tuple $(N, R, T, B)$, select a subset of observation variables $E \subseteq \mathbf{Y}$ which maximizes the expected reward:*

$$\hat{R}(\mathbf{X}_H|E) = \sum_{e \in Dom(E)} \Pr(E = e) R(\mathbf{X}_H|E = e) \quad (7)$$

*subject to the budget limit $T(E) \le B$.*

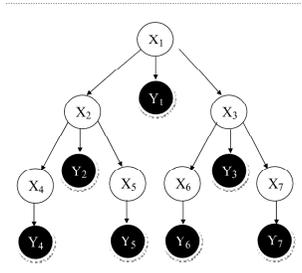

Figure 2: Out-tree BN topology.

We assume BNs with an out-tree shaped dependency graph rooted in node $X_1$ (as shown in figure 2). Let $\mathbf{X}_s^I$ denote a subset of $\mathbf{X}$, which consists of all nodes $X_i$ in the subtree rooted in $X_s$ (all descendants of $X_s$ including $X_s$ itself), and let $E \subseteq \mathbf{Y}$ be any given subset of observation ("evidence") nodes. We use the following notation to refer to other relevant subsets of nodes:

$$\mathbf{X}_s^O = \mathbf{X} \setminus \mathbf{X}_s^I;$$
$$\mathbf{Y}_s^I = \{Y_i : X_i \in (\mathbf{X}_M \cap \mathbf{X}_s^I)\};$$
$$\mathbf{Y}_s^O = \mathbf{Y} \setminus \mathbf{Y}_s^I;$$
$$E_s^I = E \cap \mathbf{Y}_s^I;$$
$$E_s^O = E \cap \mathbf{Y}_s^O.$$

### 4.1 OSS in discrete BNs

We now apply our approach to OSS in tree-shaped BNs with discrete variables. Since solving this problem for the general setting (even for the tree-structured topology) is proved to be $NP^{PP}$-hard [7], we restrict our focus to BNs with boolean variables ($Dom(X_i) = Dom(Y_i) = \{0, 1\}$), and consider the following reward function, defined for an arbitrary set of nodes $A$:

$$R(A|E = e) = \max_{X_i \in A} R_i(X_i|E = e),$$

where each $R_i : \Pr(X_i) \to \mathcal{R}$ is a local reward function:

$$R_i(X_i|E = e) = \begin{cases} \Pr(X_i = 1|E = e) & : \text{if}(X_i \in X_H), \\ 0 & : \text{otherwise.} \end{cases} \quad (8)$$

We refer to this version of OSS as *Boolean OSS* (BOSS).

The BN can be specified by the following parameters:

$$\alpha_i = \begin{cases} \Pr(X_i = 1) & : \text{if } X_i \text{ is a root}, \\ \Pr(X_i = 1|X_{Prev(i)} = 1) & : \text{otherwise}; \end{cases}$$

$$\beta_i = \begin{cases} \Pr(X_i = 1) & : \text{if } X_i \text{ is a root}, \\ \Pr(X_i = 1|X_{Prev(i)} = 0) & : \text{otherwise}; \end{cases}$$

$$\theta_i = \begin{cases} \Pr(Y_i = 0|X_i = 1) & : \text{if}(X_i \in \mathbf{X}_M), \\ 1 & : \text{otherwise}; \end{cases}$$

$$\zeta_i = \begin{cases} \Pr(Y_i = 1|X_i = 0) & : \text{if}(X_i \in \mathbf{X}_M), \\ 0 & : \text{otherwise}; \end{cases}$$

We make the following simplifying assumptions about the involved observation process:

1. Probability of a *false positive* observation result for all nodes is bounded by a small constant $\zeta_{max}$ (that is $\forall i : \zeta_i \le \zeta_{max}$);

2. Only hypothesis nodes have directly attached observations ($\mathbf{X}_M \subseteq \mathbf{X}_H$).

Henceforth this set of assumptions is called the *restricted false positive* property. In the extreme case ($\zeta_{max} = 0$) we get a *false-positive-free* observation process. Despite the relatively restricted setting, we have the following complexity result:

**Theorem 1** (Hardness of BOSS). *Finding an exact solution to the BOSS problem is NP-hard even when all state variables are independent ($\alpha_i = \beta_i$) and all observations are exact ($\theta_i = \zeta_i = 0$).*

Proof is by reduction from *Knapsack*, which is a well-known *NP*-complete problem. Below, we show how the BOSS problem can be reduced to a special case of MRA and then solved (approximately) by the RLC algorithm.

In order to apply the RLC algorithm we must specify the problem in terms of a composite system. Deriving the corresponding CG is straightforward: the graph has in-tree form and can be obtained from the dependency graph by simply reversing directions of all arcs.

Careful inspection of the *false-positive-free* property yields that observing one positive value at any observation node ($Y_i = 1$) provides a sufficient evidence for determining the reward value ($R(\mathbf{X}_H | Y_i = 1, E = e) = R_i(X_i | Y_i = 1) = 1$), regardless of other observations. We employ this fact to obtain a recursive decomposition of the expected reward.

We specify output quality ($q$-component) of exploring subset $E \subseteq \mathbf{Y}$ w.r.t. subtree $\mathbf{X}_s^I$ as the triple $(f, g, r)$:

$$f = \Pr(\hat{e}_s^I | X_s = 1),$$
$$g = \Pr(\hat{e}_s^I | X_s = 0),$$
$$r = R(\mathbf{X}_s^I | \hat{e}),$$

where $\hat{e}$, $\hat{e}_s^I$, and $\hat{e}_s^O$ denote assignments of all zeros to $E$, $E_s^I$, and $E_s^O$ respectively. While $f$ and $g$ components depend only on observations inside the subtree ($\hat{e}_s^I$), to determine the value of the $r$-component we need additional information from outside the subtree, provided by the $p$-component: $p = \Pr(X_s = 1 | \hat{e}_s^O)$.

For each quality component we define one corresponding domain set (of relevant values):

$$\mathbf{P}_s = \{\Pr(X_s = 1 | E_s^O = \hat{e}_s^O) : E \subseteq \mathbf{Y}\}$$
$$\mathbf{F}_s = \{\Pr(E_s^I = \hat{e}_s^I | X_s = 1) : E \subseteq \mathbf{Y}\}$$
$$\mathbf{G}_s = \{\Pr(E_s^I = \hat{e}_s^I | X_s = 0) : E \subseteq \mathbf{Y}\}$$
$$\mathbf{R}_s = \{R(\mathbf{X}_s^I | E = \hat{e}) : E \subseteq \mathbf{Y}\}.$$

We also define combined domain sets $\mathbf{Q}_s = \mathbf{F}_s \times \mathbf{G}_s \times \mathbf{R}_s$. Finally, all alternative assignments to a number of observations (measurements) in node $X_s$ is represented by set $\mathbf{M}_s$. In the basic OSS setting we have at most one observation per node:

$$\mathbf{M}_s = \begin{cases} \{0, 1\} & : \text{if } (X_s \in \mathbf{X}_M), \\ \{0\} & : \text{otherwise.} \end{cases}$$

However, the model can be easily extended to multiple observations (by specifying appropriate $\mathbf{M}_s$ sets).

RPPs of observation nodes contain CPs with no condition (denoted by $\emptyset$ in the $p$-component):

$$Q_s^Y = \{(m\hat{\tau}_s, \emptyset, m) : m \in \mathbf{M}_s\}, \qquad (9)$$

where $\hat{\tau}_s$ is an assignment of $\tau_s$ time units to $s$ and 0 to all the other CCs.

All leaf $X$-nodes are associated with RPPs of the following form:

$$Q_s^X = \{(\hat{0}, u, \psi_s(u)) : u \in \mathbf{P}_s \times \mathbf{M}_s\} \qquad (10)$$

where $\hat{0}$ denotes a zero time allocation (to all CCs), and $\psi_s : \mathbf{P}_s \times \mathbf{M}_s \to \mathbf{Q}_s$ is a vector function defined as follows:

$$\psi_s(p, m) = (f, g, r), \qquad (11)$$
$$f = \Pr(\hat{e}_s^I | X_s = 1) = \theta_s^m,$$
$$g = \Pr(\hat{e}_s^I | X_s = 0) = (1 - \zeta_s)^m,$$
$$r = R_s(X_s | \hat{e}) = \begin{cases} \frac{pf}{L(f,g,p)} & : \text{if } (X_s \in \mathbf{X}_H), \\ 0 & : \text{otherwise} \end{cases}$$

In our notation $L(\cdot, \cdot, \cdot)$ stands for the operator of linear interpolation defined as follows:

$$L(a, b, c) = ca + (1 - c)b. \qquad (12)$$

RPP of any non-leaf node $X_s$ with $k$ children $(X_{s_1}, \ldots, X_{s_k})$ is specified as follows:

$$Q_s^X = \{(\hat{0}, u, \psi_s(u)) : u \in \mathbf{P}_s \times \mathbf{M}_s \times \mathbf{Q}_{s_1} \times \cdots \times \mathbf{Q}_{s_k}\} \qquad (13)$$

where
$\psi_s : \mathbf{P}_s \times \mathbf{M}_s \times \mathbf{Q}_{s_1} \times \cdots \times \mathbf{Q}_{s_k} \to \mathbf{Q}_s \times \mathbf{P}_{s_1} \times \cdots \times \mathbf{P}_{s_k}$
is a vector function defined as follows:

$$\psi_s(p, m, (f_1, g_1, r_1), \ldots, (f_k, g_k, r_k)) = \qquad (14)$$
$$= ((f, g, r), p_1, \ldots, p_k),$$
$$r = \max\{r_0, r_1, \ldots, r_k\},$$
$$r_0 = R_s(X_s | \hat{e}) = \begin{cases} \frac{pf}{L(f,g,p)} & : \text{if } (X_s \in \mathbf{X}_H), \\ 0 & : \text{otherwise} \end{cases}$$
$$f = \Pr(\hat{e}_s^I | X_s = 1) = \theta_s^m \prod_{1 \leq i \leq k} f_i',$$
$$g = \Pr(\hat{e}_s^I | X_s = 0) = (1 - \zeta_s)^m \prod_{1 \leq i \leq k} g_i',$$
$$p_i = \Pr(X_{s_i} = 1 | \hat{e}_{s_i}^O) =$$
$$= \frac{L(\alpha_{s_i}, \beta_{s_i}, p\theta_s^m \prod_{j \neq i} f_j')}{L(\theta_s^m \prod_{j \neq i} f_j', (1 - \zeta_s)^m \prod_{j \neq i} g_j', p)},$$
$$f_i' = \Pr(\hat{e}_{s_i}^I | X_s = 1) = L(f_i, g_i, \alpha_{s_i}),$$
$$g_i' = \Pr(\hat{e}_{s_i}^I | X_s = 0) = L(f_i, g_i, \beta_{s_i}).$$

While all $Q_s^Y$ RPPs can be specified explicitly (as a list of CPs), the $Q_s^X$ RPPs generally cannot, due to a possibly exponential size of domains $\mathbf{P}_s$, $\mathbf{F}_s$, $\mathbf{G}_s$ and $\mathbf{R}_s$ in their input condition. Instead, we represent them in an implicit form by providing (parameters of) the involved $\psi_s$ functions.

During the compilation process (according to the RLC algorithm) method *Compose* is applied to lists of RPPs. At each application one composed RPP $Q_s^c$ (which represents the whole subtree $\mathbf{X}_s^I$) is generated. Due to backward conditioning on elements of $\mathbf{P}_s$ which can contain exponential number of elements), such a RPP cannot be derived exactly (and explicitly). There is also no obvious way to specify it implicitly by providing some predefined functions (as we do in case of individual RPPs). To address this problem, we propose to approximate all $\mathbf{P}_s$ sets by one uniform grid $\mathbf{D}_p$ defined as follows:

$$\mathbf{D}_p = \left\{ \left(k + \frac{1}{2}\right) \epsilon_p : k \in 0, \ldots, d_p - 1 \right\} \quad (15)$$

where $0 < \epsilon_p < 1$ is a small constant, and $d_p = \left\lceil \frac{1}{\epsilon_p} \right\rceil$ is a number of intervals of size $\epsilon_p$ in the range $[0, 1]$. Thus, set $\mathbf{D}_p$ has a fixed number of elements ($d_p$) which makes efficient enumeration possible.

We define discretization function $\lambda_p$, which maps any value $p \in \mathbf{P}_s$ to an appropriate point in set $\mathbf{D}_p$:

$$\lambda_p(p) = \left( \left\lfloor \frac{p}{\epsilon_p} \right\rfloor + \frac{1}{2} \right) \epsilon_p \quad (16)$$

This discretization induces approximate equivalence relation $\stackrel{P}{\approx}$ among elements of $\mathbf{P}_s$ defined as follows:

$$p \stackrel{P}{\approx} p' \Leftrightarrow \lambda_p(p) = \lambda_p(p') \quad (17)$$

We apply a similar discretization technique to other domain sets ($\mathbf{F}_s$, $\mathbf{G}_s$, and $\mathbf{R}_s$) with discretization steps $\epsilon_f$, $\epsilon_g$ and $\epsilon_r$ respectively. The appropriate grids ($\mathbf{D}_f$, $\mathbf{D}_g$, $\mathbf{D}_r$), discretization functions ($\lambda_f$, $\lambda_g$, $\lambda_r$) and *equivalence* operators ($\stackrel{F}{\approx}$, $\stackrel{G}{\approx}$, $\stackrel{R}{\approx}$) are defined accordingly. The composed *equivalence* operator $\stackrel{Q}{\approx}$ is defined as follows:

$$(f, g, r) \stackrel{Q}{\approx} (f', g', r') \Leftrightarrow (f \stackrel{F}{\approx} f') \wedge (g \stackrel{G}{\approx} g') \wedge (r \stackrel{R}{\approx} r')$$

To complete the specification, we define the following comprehensive utility function:

$$U(\hat{t}, p, (f, g, r)) = \quad (18)$$
$$= \begin{cases} L(r, 1, L(f, g, \alpha_1)) & : \text{if } (p \stackrel{P}{\approx} \alpha_1 \wedge \|\hat{t}\| \leq B), \\ -\infty & : \text{otherwise,} \end{cases}$$

After compiling RPPs of the entire tree, the resulting RPP $Q_1^c$ can be used to select a near-optimal TAS w.r.t. this utility function. Let $E^*$ denote the optimal observation subset, and let $E$ be a subset corresponding to the TAS selected based on the resulting RPP.

**Theorem 2** (Approximation quality of RLC for BOSS). *The RLC algorithm applied to a transformed instance of BOSS with out-tree topology and a restricted false positive observation process approximates the optimal solution within additive factor of $\Delta_u$, which is bounded as follows:*

$$\Delta_u = \hat{R}(\mathbf{X}|E^*) - \hat{R}(\mathbf{X}|E) \leq h\epsilon_p + 2n\epsilon_s + \epsilon_r + n\zeta_{max} \quad (19)$$

*where $h$ is a height of the tree, and $\epsilon_s = \max\{\epsilon_f, \epsilon_g\}$.*

Selection of the appropriate values for the grid steps depends on the required precision of the solution. To ensure approximation with $\Delta_u \leq \epsilon + n\zeta_{max}$ (in worst case) we can select $\epsilon_p = \frac{\epsilon}{3h}$, $\epsilon_f = \epsilon_g = \frac{\epsilon}{6n}$, and $\epsilon_r = \frac{\epsilon}{3}$.

Due to monotonicity in total time, any composed RPP can be represented by the 4-dimensional table ($\mathbf{D}_p \times \mathbf{D}_f \times \mathbf{D}_g \times \mathbf{D}_r$), where each entry is associated with at most one appropriate partial TAS. The number of entries (CPs) in such a table is bounded as follows:

$$|Q_s^c| \leq d_p d_f d_g d_r = \left\lceil \frac{3h}{\epsilon} \right\rceil \left\lceil \frac{6n}{\epsilon} \right\rceil^2 \left\lceil \frac{3}{\epsilon} \right\rceil. \quad (20)$$

Worst-case complexity for the complete run of RLC is $O\left(\frac{n^4}{\epsilon^4}\right)$ time, $O\left(\frac{n^3}{\epsilon^4}\right)$ space for a chain topology, and $O\left(\frac{n^{2c+1}hc}{\epsilon^{2c+2}}\right)$ time, $O\left(\frac{n^{2c}h^2c}{\epsilon^{2c+2}}\right)$ space for a tree with a maximum branching factor of $c$.

### 4.2 OSS in Gaussian Bayesian networks

*Gaussian Bayesian network* (GBN) is a special case of BN, where the *conditional probability distributions* of the variables are Normal (Gaussian) distributions:

$$X_i | Pa(X_i) \sim N\left(\mu_i + \sum_{X_j \in Pa(X_i)} a_{i,j}(X_j - \mu_j), \sigma_i^2\right)$$

We parametrize our GBN model as follows:

$$\alpha_i = a_{i, Prev(i)}^2,$$
$$\beta_i = Prec(X_i | X_{Prev(i)}) = \frac{1}{\sigma_i^2},$$
$$\theta_i = \begin{cases} Prec(Y_i | X_i) & : \text{if } (X_i \in \mathbf{X}_M), \\ 0 & : \text{otherwise.} \end{cases}$$

In our notation $Prec(\cdot)$ denotes the precision operator, which is reciprocal to variance:

$$Prec(X_i | E) = \frac{1}{Var(X_i | E)}. \quad (21)$$

For *Gaussian OSS* (GOSS) we consider the following reward function:

$$R(A|E=e) = \min_{X_i \in A} R_i(X_i|E),$$

where $R_i : \Pr(X_i) \to \mathcal{R}$ are local reward functions:

$$R_i(X_i|E) = \begin{cases} \overline{\log}_{a,b}(Prec(X_i|E)) & : \text{if}(X_i \in \mathbf{X}_H), \\ 1 & : \text{otherwise.} \end{cases} \quad (22)$$

Here $a$ and $b$ are two parameters that determine a range of distinguishable (for reward) values of precision, and $\overline{\log}_{a,b}(\cdot)$ denotes a normalized (and truncated at its extreme points) log operator, defined as follows:

$$\overline{\log}_{a,b}(p) = \begin{cases} 0 & : \text{if } (p \leq a), \\ 1 & : \text{if } (p \geq b), \\ \frac{\log p - \log a}{\log b - \log a} & : \text{otherwise.} \end{cases} \quad (23)$$

**Theorem 3** (Hardness of GOSS). *Finding an exact solution for a general instance of the GOSS problem is NP-hard even for a tree-shaped GBN.*

Proof is by reduction from *Knapsack*. A polynomial scheme for approximate solution of GOSS, similar to one presented for BOSS, follows.

To apply the RLC algorithm we specify the problem in terms of a composite system. The CG is as for BOSS. All domain sets (except for $\mathbf{M}_s$, that remain the same) should be redefined as follows:

$$\mathbf{P}_s = \{Prec(X_s|E_s^O) : E \subseteq \mathbf{Y}\}, \quad (24)$$
$$\mathbf{F}_s = \{Prec(X_s|E_s^I) : E \subseteq \mathbf{Y}\}, \quad (25)$$
$$\mathbf{R}_s = \{R_s(X_s|E) : E \subseteq \mathbf{Y}\}, \quad (26)$$
$$\mathbf{Q}_s = \mathbf{F}_s \times \mathbf{R}_s. \quad (27)$$

We need to reformulate, in the definition of RPPs, the specification of the $\psi_s$ functions. For the leaf nodes $\psi_s$ is defined as follows:

$$\psi_s(p, m) = (f, r), \quad (28)$$
$$f = Prec(X_s|E_s^I) = m\theta_s,$$
$$r = R_s(X_s|E) = \begin{cases} \overline{\log}_{a,b}(p+f) & : \text{if } (X_s \in \mathbf{X}_H), \\ 1 & : \text{otherwise,} \end{cases}$$

For the non-leaf nodes we have:

$$\psi_s(p, m, (f_1, r_1), \ldots, (f_k, r_k)) = ((f, r), p_1, \ldots, p_k);$$
$$f = Prec(X_s|E_s^I) = m\theta_s + \sum_{1 \leq i \leq k} f_i',$$
$$r = \min\{r_0, r_1, \ldots, r_k\},$$
$$r_0 = R_s(X_s|E) = \begin{cases} \overline{\log}_{a,b}(p+f) & : \text{if } (X_s \in \mathbf{X}_H), \\ 1 & : \text{otherwise,} \end{cases}$$
$$p_i = Prec(X_{s_i}|E_{s_i}^O) = J(\beta_{s_i}, \alpha_{s_i}(m\theta_s + p + \sum_{j \neq i} f_j')),$$
$$f_i' = Prec(X_s|E_{s_i}^I) = \frac{J(f_i, \beta_{s_i})}{\alpha_{s_i}}, \quad (29)$$

$$(30)$$

In our notation $J(\cdot, \cdot)$ stands for the operator of precision propagation defined as follows:

$$J(a, b) = \begin{cases} 0 & : \text{if } (a = b = 0), \\ \frac{ab}{a+b} & : \text{otherwise.} \end{cases} \quad (31)$$

As in case of BOSS, to prevent exponential growth of the composed RPPs we apply discretization to all domains by appropriate grids. Grid $\mathbf{D}_r$ and corresponding discretization function $\lambda_r$ are defined exactly as in BOSS. To define $\mathbf{D}_p$ we use its projection $\mathbf{D}_p'$ to the $[0, 1]$ interval ($\mathbf{D}_p'$ is defined exactly as $\mathbf{D}_p$ in the BOSS case):

$$\mathbf{D}_p = \{p : \overline{\log}_{a,b}(p) \in \mathbf{D}_p'\}, \quad (32)$$

We express the discretization function $\lambda_p$ through its projected version $\lambda_p'$ (which is defined as $\lambda_p$ in BOSS):

$$\lambda_p(p) = \lambda_p'(\overline{\log}_{a,b}(p)). \quad (33)$$

Grid $\mathbf{D}_f$ and the corresponding discretization function $\lambda_f$ are similarly defined.

*Equivalence* operators $\stackrel{P}{\approx}$, $\stackrel{F}{\approx}$, and $\stackrel{R}{\approx}$ are defined as in BOSS. The composed *equivalence* operator $\stackrel{Q}{\approx}$ is:

$$(f, r) \stackrel{Q}{\approx} (f', r') \Leftrightarrow (f \stackrel{F}{\approx} f') \wedge (r \stackrel{R}{\approx} r')$$

The comprehensive utility function is as follows:

$$U(\hat{t}, p, (f, r)) = \begin{cases} r & : \text{if } ((p \stackrel{P}{\approx} \beta_1) \wedge (\|\hat{t}\| \leq B)), \\ -\infty & : \text{otherwise} \end{cases}$$

$$(34)$$

After compiling the composite system (using the RLC algorithm), a near-optimal TAS can be selected from the resulting RPP $Q_1^c$ w.r.t. this utility function. Let $E^*$ denote an optimal observation subset, and $E$ a subset corresponding to the TAS selected from $Q_1^c$.

**Theorem 4** (Approximation quality of RLC for GOSS). *The RLC algorithm applied to a transformed instance of GOSS problem with out-tree topology approximates the optimal solution $E^*$ within additive factor of $\Delta_u$, bounded as follows:*

$$\Delta_u = \hat{R}(\mathbf{X}_H|E^*) - \hat{R}(\mathbf{X}_H|E) \leq h\epsilon_p + h\epsilon_f + \epsilon_r \quad (35)$$

To ensure approximation with $\Delta_u \leq \epsilon$ (in worst case) we can select $\epsilon_p = \epsilon_f = \frac{\epsilon}{h}$, and $\epsilon_r = \epsilon$.

Any composed RPP $Q_s^c$ can be represented by a 3-dimensional $(\mathbf{D}_p \times \mathbf{D}_f \times \mathbf{D}_r)$ table with a number of entries bounded as follows:

$$|Q_s^c| \leq d_p d_f d_r = \left\lceil\frac{h}{\epsilon}\right\rceil^2 \left\lceil\frac{1}{\epsilon}\right\rceil. \quad (36)$$

The appropriate worst-case complexity for the complete run of the RLC algorithm is $O\left(\frac{nh^2}{\epsilon^2}\right)$ time, $O\left(\frac{h^2}{\epsilon^3}\right)$ space for a chain topology, and $O\left(\frac{nh^{c+1}c}{\epsilon^{c+2}}\right)$ time, $O\left(\frac{h^{c+2}c}{\epsilon^{c+2}}\right)$ space for a tree with maximum branching factor of $c$.

## 5 Summary

In this paper we extended the concept of CPP, and presented an efficient technique for compiling a composite system beyond the *input monotonicity* assumption. The extended scheme has been applied to optimizing a set of measurements in two different settings (for choosing a maximum expectation variable in a binary valued BN, and for minimizing the worst variance in a Gaussian BN). Polynomial time methods have been presented for both problems, and quality of approximation has been theoretically determined.

Applying our framework to real-world domains as an empirical evaluation is underway. Further extending the framework to deal with more general system topologies, tractable strategies for active monitoring are possible directions for future research.

#### Acknowledgements

Partially supported by the IMG4 consortium (under the Ministry of Industry, Trade and Labor of Israel MAGNET program), by the Lynn and William Frankel Center for Computer Sciences, and by the Paul Ivanier Center for Robotics.

## References

[1] M. Boddy and T. Dean. Solving time-dependent planning problems. In *IJCAI*, pages 979–984, 1989.

[2] A. Cassandra, M. L. Littman, and N. L. Zhang. Incremental Pruning: A simple, fast, exact method for partially observable Markov decision processes. In *Proceedings of (UAI–97)*, pages 54–61, 1997.

[3] E. Horvitz. Reasoning about beliefs and actions under computational resource constraints. In *UAI*, pages 301–324, 1987.

[4] E. Horvitz. Models of continual computation. In *AAAI/IAAI*, pages 286–293, 1997.

[5] E. Horvitz. Continual computation policies for allocating offline and real-time resources. In *IJCAI*, pages 1280–1287, 1999.

[6] A. Krause and C. Guestrin. Near-optimal nonmyopic value of information in graphical models. In *UAI*, pages 324–331, 2005.

[7] A. Krause and C. Guestrin. Optimal nonmyopic value of information in graphical models - efficient algorithms and theoretical limits. In *IJCAI*, pages 1339–1345, 2005.

[8] A. I. Mouaddib and S. Zilberstein. Knowledge-based anytime computation. In *IJCAI*, pages 775–783, 1995.

[9] Y. Radovilsky, G. Shattah, and E. S. Shimony. Efficient deterministic approximation algorithm for nonmyopic value of information in graphical models. In *SMC Conference*, Taipei, Taiwan, 2006.

[10] S. Zilberstein. Operational rationality through compilation of anytime algorithms. Technical report, Computer Science Division, University of California at Berkeley, 1993. PhD Dissertation.

[11] S. Zilberstein. Optimizing decision quality with contract algorithms. In *IJCAI*, pages 1576–1582, 1995.

[12] S. Zilberstein. Using anytime algorithms in intelligent systems. *AI Magazine*, 17(3):73–83, 1996.